\documentstyle[aclap]{article}
\title{Creating a tagset, lexicon and guesser for a French tagger\thanks{
presented in the ACL SIGDAT workshop on {\em From Texts To Tags:
Issues In Multilingual Language Analysis}.  pages 58-64. University
College Dublin, Ireland, 1995.}}
\author{Jean-Pierre Chanod \and Pasi Tapanainen \\
Rank Xerox Research Centre, Grenoble Laboratory \\
6, chemin de Maupertuis, 38240 Meylan, France \\
{\tt Jean.Pierre.Chanod,Pasi.Tapanainen@xerox.fr} \\
{\tt http://www.xerox.fr/grenoble/mltt/home.html}
}

\newenvironment{acknowledgments}{\small\par{\bf\noindent
Acknowledgments}\par\vspace{2pt}\raggedright}{\par}

\begin{document}
\maketitle

\begin{abstract}
We earlier described two taggers for French, a statistical one and a
constraint-based one.  The two taggers have the same tokeniser and
morphological analyser.  In this paper, we describe aspects of this
work concerned with the definition of the tagset, the building of the
lexicon, derived from an existing two-level morphological analyser,
and the definition of a lexical transducer for guessing unknown words.
\end{abstract}

\section{Background}

We earlier described two taggers for French: the statistical one
having an accuracy of 95--97~\% and the constraint-based one 97--99~\%
(see \cite{CT94,CT95}).  The disambiguation has been already
described, and here we discuss the other stages of the process, namely
the definition of the tagset, transforming a current lexicon into a
new one and guessing the words that do not appear in the lexicon.

Our lexicon is based on a finite-state transducer lexicon
\cite{KKZ92}.  The French description was originally built by Annie
Zaenen and Carol Neidle, and later refined by Jean-Pierre Chanod
\shortcite{Ch94}.

Related work on French can be found in \cite{AMD85}.

\section{Tagset}

We describe in this section criteria for selecting the tagset.  The
following is based on what we noticed to be useful during the
developing the taggers.

\subsection{The size of the tagset}

Our basic French morphological analyser was not originally designed
for a (statistical) tagger and the number of different tag
combinations it has is quite high.  The size of the tagset is only 88.
But because a word is typically associated with a sequence of tags,
the number of different combinations is higher, 353 possible sequences
for single French words.  If we also consider words joined with
clitics, the number of different combinations is much higher, namely
6525.

A big tagset does not cause trouble for a constraint-based tagger
because one can refer to a combination of tags as easily as to a
single tag.  For a statistical tagger however, a big tagset may be a
major problem.  We therefore used two principles for forming the
tagset: (1) the tagset should not be big and (2) the tagset should not
introduce distinctions that cannot be resolved at this level of
analysis.

\subsection{Verb tense and mood}

As distinctions that cannot be resolved at this level of analysis
should be avoided, we do not have information about the tense of the
verbs.  Some of this information can be recovered later by performing
another lexicon lookup after the analysis.  Thus, if the verb tense is
not ambiguous, we have not lost any information and, even if it is, a
part-of-speech tagger could not resolve the ambiguity very reliably
anyway.  For instance, {\em dort} (present; {\em sleeps}) and {\em
dormira} (future; {\em will sleep}) have the same tag {\em
VERB-SG-P3}, because they are both singular, third-person forms and
they can both be the main verb of a clause.  If needed, we can do
another lexicon lookup for words that have the tag {\em VERB-SG-P3}
and assign a tense to them after the disambiguation.  Therefore, the
tagset and the lexicon together may make finer distinctions than the
tagger alone.

On the other hand, the same verb form {\em dit} can be third person
singular present indicative or third person singular past historic
(pass\'{e} simple) of the verb {\em dire} ({\em to say}).  We do not
introduce the distinction between those two forms, both tagged as {\em
VERB-SG-P3}, because determining which of the two tenses is to be
selected in a given context goes beyond the scope of the tagger.
However, we do keep the distinction between {\em dit} as a finite verb
(present or past) on one side and as a past participle on the other,
because this distinction is properly handled with a limited contextual
analysis.

Morphological information concerning mood is also collapsed in the
same way, so that a large class of ambiguity between present
indicative and present subjunctive is not resolved: again this is
motivated by the fact that the mood is determined by remote elements
such as, among others, connectors that can be located at
(theoretically) any distance from the verb.  For instance, a
conjunction like {\em quoique} requires the subjunctive mood:
\begin{quote}
Quoique, en principe, ce cas {\bf soit} fr\'{e}quent.
(Though, in principle, this case {\bf is} [subjunctive] frequent.)
\end{quote}

The polarity of the main verb to which a subordinate clause is
attached also plays a role.  For instance, compare:
\begin{quote}
Je pense que les petits enfants {\bf font} de jolis dessins.
(I think that small kids {\bf make} [indicative] nice drawings.) \\
\\
Je ne pense pas pas que les petits enfants {\bf fassent} de jolis dessins.
(I do not think that small kids {\bf make} [subjunctive]  nice drawings.) \\
\end{quote}
Consequently, forms like {\em chante} are tagged as VERB-P3SG
regardless of their mood.  In the case of {\em faire} (to do, to make)
however, the mood information can easily be recovered as the third
person plural are {\em font} and {\em fassent} for indicative and
subjunctive moods respectively.

\subsection{Person}

The person seems to be problematic for a statistical tagger (but not
for a constraint-based tagger).  For instance, the verb {\em pense},
ambiguous between the first- and third-person, in the sentence {\em Je
ne le pense pas} (I do not think so) is disambiguated wrongly because
the statistical tagger fails to see the first-person pronoun {\em je}
and selects more common third-person reading for the verb.

We made a choice to collapse the first- and second-person verbs
together but not the third person.  The reason why we cannot also
collapse the third person is that we have an ambiguity class that
contains adjective and first- or second-person verbs.  In a sentence
like {\em Le secteur mati\`{e}res (NOUN-PL) plastiques
(ADJ-PL/NOUN-PL/VERB-P1P2)\ldots} the verb reading for {\em
plastiques} is impossible.  Because noun --- third-person sequence is
relatively common, collapsing also the third person would cause
trouble in parsing.

Because we use the same tag for first- and second-person verbs, the
first- and second-person pronouns are also collapsed together to keep
the system consistent.  Determining the person after the analysis is
also quite straightforward: the personal pronouns are not ambiguous,
and the verb form, if it is ambiguous, can be recovered from its
subject pronoun.

\subsection{Lexical word-form}

Surface forms under a same lexical item were also collapsed when they
can be attached to different lemmata (lexical forms) while sharing the
same category, such as {\em peignent} derived from the verb {\em
peigner} ({\em to comb}) or {\em peindre} ({\em to paint}).  Such
coincidental situations are very rare in French \cite{Elb93}.
However, in the case of {\em suis} first person singular of the
auxiliary {\em \^{e}tre} ({\em to be}) or of the verb {\em suivre}
({\em to follow}), the distinction is maintained, as we introduced
special tags for auxiliaries.

\subsection{Gender and number}

We have not introduced gender distinctions as far as nouns and
adjectives (and incidentally determiners) are concerned.  Thus a
feminine noun like {\em chaise} ({\em chair}) and a masculine noun
like {\em tabouret} ({\em stool}) both receive the same tag {\em
NOUN-SG}.

However, we have introduced distinctions between singular nouns ({\em
NOUN-SG}), plural nouns ({\em NOUN-PL}) and number-invariant nouns
({\em NOUN-INV}) such as {\em taux} ({\em rate/rates}).  Similar
distinctions apply to adjectives and determiners.  The main reason for
this choice is that number, unlike gender, plays a major role in
French with respect to subject/verb agreement, and the noun/verb
ambiguity is one of the major cases that we want the tagger to
resolve.

\subsection{Discussion on Gender}

Ignoring gender distinction for a French tagger is certainly counter
intuitive.  There are three major objections against this choice:
\begin{itemize}
\item Gender information would provide better disambiguation,
\item Gender ambiguous nouns should be resolved, and
\item Displaying gender provides more information.
\end{itemize}


There is obviously a strong objection against leaving out gender
information as this information may provide a better disambiguation in
some contexts.  For instance in {\em le diffuseur diffuse}, the word
{\em diffuse} is ambiguous as a verb or as a feminine adjective.  This
last category is unlikely after a masculine noun like {\em diffuseur}.

However, one may observe that gender agreement between nouns and
adjectives often involve long distance dependencies, due for instance
to coordination or to the adjunction of noun complements as in {\em
une envie de soleil diffuse} where the feminine adjective {\em
diffuse} agrees with the feminine noun {\em envie}.  In other words,
introducing linguistically relevant information such as gender into
the tagset is fine, but if this information is not used in the
linguistically relevant context, the benefit is unclear.  Therefore,
if a (statistical) tagger is not able to use the relevant context, it
may produce some extra errors by using the gender.

An interesting, albeit minor interest of not introducing gender
distinction, is that there is then no problem with tagging phrases
like {\em mon allusion} ({\em my allusion}) where the masculine form
of the possessive determiner {\em mon} precedes a feminine singular
noun that begins with a vowel, for euphonic reasons.

Our position is that situations where the gender distinction would
help are rare, and that the expected improvement could well be
impaired by new errors in some other contexts.  On a test suite
\cite{CT95} extracted from the newspaper Le Monde (12~000 words)
tagged with either of our two taggers, we counted only three errors
that violated gender agreement.  Two could have been avoided by other
means, i.e.~they belong to other classes of tagging errors.  The
problematic sentence was:
\begin{quote}
 L'arm\'{e}e interdit d'autre part le passage\ldots\\
(The army forbids the passage\ldots)
\end{quote}
where {\em interdit} is mistakenly tagged as an adjective rather than
a finite verb, while {\em arm\'{e}e} is a feminine noun and {\em
interdit} a masculine adjective, which makes the {\em noun--adjective}
sequence impossible in this particular sentence\footnote{We have not
systematically compared the two approaches, i.e.~with or without
gender distinction, but previous experiences \cite{Ch93} with broad
coverage parsing of possibly erroneous texts have shown that gender
agreement is not as essential as one may think when it comes to French
parsing.}.


Another argument in favour of gender distinction is that some nouns
are ambiguously masculine or feminine, with possible differences in
meaning, e.g.~{\em poste}, {\em garde}, {\em manche}, {\em tour}, {\em
page}.  A tagger that would carry on the distinction would then
provide sense disambiguation for such words.

Actually, such gender-ambiguous words are not very frequent.  On the
same 12~000-word test corpus, we counted 46 occurrences of words which
have different meanings for the masculine and the feminine noun
readings.  This number could be further reduced if extremely rare
readings were removed from the lexicon, like masculine {\em ombre} (a
kind of fish while the feminine reading means shadow or shade) or
feminine {\em litre} (a religious ornament).  We also counted 325
occurrences of nouns (proper nouns excluded) which do not have
different meanings in the masculine and the feminine readings,
e.g.~{\em \'{e}l\`{e}ve}, {\em camarade}, {\em jeune}.

A reason not to distinguish the gender of such nouns, besides their
sparsity, is that the immediate context does not always suffice to
resolve the ambiguity.  Basically, disambiguation is possible if there
is an unambiguous masculine or feminine modifier attached to the noun
as in {\em le poste} vs.~{\em la poste}.  This is often not the case,
especially for {\em preposition + noun} sequences and for plural
forms, as plural determiners themselves are often ambiguous with
respect to gender.  For instance, in our test corpus, we find
expressions like {\em en 225 pages}, {\em\`{a} leur tour}, {\em\`{a}
ces postes} and {\em pour les postes de responsabilit\'{e}} for which
the contextual analysis does not help to disambiguate the gender of
the head noun.


Finally, carrying the gender information does not itself increase the
disambiguation power of the tagger.  A disambiguator that would
explicitly mark gender distinctions in the tagset would not
necessarily provide more information.  A reasonable way to assess the
disambiguating power of a tagger is to consider the ratio between the
initial number of ambiguous tags vs.~the final number of tags after
disambiguation.  For instance, it does not make any difference if the
ambiguity class for a word like {\em table} is {\em [feminine-noun,
finite-verb]} or {\em [noun, finite-verb]}, in both cases the tagger
reduces the ambiguity by a ratio of 2 to 1.  The information that can
be derived from this disambiguation is a matter of associating the
tagged word with any relevant information like its base form,
morphological features such as gender, or even its definition or its
translation into some other language.  This can be achieved by looking
up the disambiguated word in the appropriate lexicon.  Providing this
derived information is not an intrinsic property of the tagger.

Our point is that the objections do not hold very strongly.  Gender
information is certainly important in itself.  We only argue that
ignoring it at the level of part-of-speech tagging has no measurable
effect on the overall quality of the tagger.  On our test corpus of
12~000 words, only three errors violate gender agreement.  This
indicates how little the accuracy of the tagger could be improved by
introducing gender distinction.  On the other hand, we do not know how
many errors would have been introduced if we had distinguished between
the genders.

\subsection{Remaining categories}

We avoid categories that are too small, i.e.~rare words that do not
fit into an existing category are collapsed together.  Making a
distinction between categories is not useful if there are not enough
occurrences of them in the training sample.  We made a category {\em
MISC} for all those miscellaneous words that do not fit into any
existing category.  This accounts for words such as: interjection {\em
oh}, salutation {\em bonjour}, onomatopoeia {\em miaou}, wordparts
i.e.~words that only exist as part of a multi-word expression, such as
{\em priori}, as part of {\em a priori}.

\subsection{Dividing a category}

In a few instances, we introduced new categories for words that have a
specific syntactic distribution.  For instance, we introduced a
word-specific tag {\em PREP-DE} for words {\em de}, {\em des} and {\em
du}, and tag {\em PREP-A} for words {\em \`{a}}, {\em au} and {\em
aux}.  Word-specific tags for other prepositions could be considered
too.  The other readings of the words were not removed, e.g.~{\em de}
is, ambiguously, still a determiner as well as {\em PREP-DE}.

When we have only one tag for all the prepositions, for example, a
sequence like
\begin{quote}
determiner noun noun/verb preposition
\end{quote}
is frequently disambiguated in the wrong way by the statistical
tagger, e.g.~{\em Le train part \`a cinq heures} ({\em The train
leaves at 5 o'clock}).  The word {\em part} is ambiguous between a
noun and a verb (singular, third person), and the tagger seems to
prefer the noun reading between a singular noun and a preposition.

We succeeded in fixing this without modifying the tagset but the
side-effect was that overall accuracy deteriorated.  The main problem
is that the preposition {\em de}, comparable to English {\em of}, is
the most common preposition and also has a specific distribution.
When we added new tags, say {\em PREP-DE} and {\em PREP-A}, for the
specific prepositions while the other prepositions remained marked
with {\em PREP}, we got the correct result, with no noticeable change
in overall accuracy.

\section{Building the lexicon}

We have a lexical transducer for French \cite{KKZ92} which was built
using Xerox Lexical Tools \cite{Xt92,Xl93}.  In our work we do not
modify the corresponding source lexicon but we employ our finite-state
calculus to map the lexical transducer into a new one.  Writing rules
that map a tag or a sequence of tags into a new tag is rather
straightforward, but redefining the source lexicon would imply complex
and time consuming work.

The initial lexicon contains all the inflectional information.  For
instance, the word {\em danses} (the plural of the noun {\em danse} or
a second person form of the verb {\em danser} ({\em to dance}) has the
following analyses\footnote{The tags represent: {\em present
indicative, singular, second person, verb}; {\em present subjunctive,
singular, second person, verb}; and {\em feminine, plural, noun}}:
\begin{verbatim}
 danser +IndP +SG +P2 +Verb
 danser +SubjP +SG +P2 +Verb
 danse  +Fem +PL +Noun
\end{verbatim}

Forms that include clitics are analysed as a sequence of items
separated by the symbols $<$ or $>$ depending on whether the clitics
precede or follow the head word.  For instance {\em vient-il} ({\em
does he come}, lit. {\em comes-he}) is analysed as\footnote{The tags
for {\em il} represent: {\em nominative, masculine, singular, third
person, clitic pronoun}.}:
\begin{verbatim}
 venir +IndP +SG +P3 +Verb
         > il +Nom +Masc +SG +P3 +PC
\end{verbatim}

 From this basic morphological transducer, we derived a new lexicon
that matches the reduced tagset described above.  This involved two
major operations:
\begin{itemize}
\item handling cliticised forms appropriately for the tagger's needs.
\item switching tagsets
\end{itemize}
In order to reduce the number of tags, cliticised items (like {\em
vient-il} are split into independent tokens for the tagging
application.  This splitting is performed at an early stage by the
tokeniser, before dictionary lookup.  Keeping track of the fact that
the tokens were initially agglutinated reduces the overall ambiguity.
For instance, if the word {\em danses} is derived from the expression
{\em danses-tu} ({\em do you dance}, lit. dance-you), then it can only
be a verb reading. This is why forms like {\em danses-tu} are
tokenised as {\em danses-} and {\em tu}, and forms like {\em
chante-t-il} are tokenised as {\em chante-t-} and {\em il}.  This in
turn requires that forms like {\em danses-} and {\em chante-t-} be
introduced into the new lexicon.

With respect to switching tagsets, we use contextual two-level rules
that turn the initial tags into new tags or to the void symbol if old
tags must simply disappear.  For instance, the symbol {\em +Verb} is
transformed into {\em +VERB-P3SG} if the immediate left context
consists of the symbols {\em +SG +P3}.  The symbols {\em +IndP}, {\em
+SG} and {\em +P3} are then transduced to the void symbol, so that
{\em vient} (or even the new token {\em vient-}) gets analysed merely
as {\em +VERB-P3SG} instead of {\em +IndP +SG +P3 +Verb}.

A final transformation consists in associating a given surface form
with its ambiguity class, i.e.~with the alphabetically ordered
sequence of all its possible tags.  For instance {\em danses} is
associated with the ambiguity class {\em [+NOUN-PL +VERB-P1P2]},
i.e.~it is either a plural noun or a verb form that belongs to the
collapsed first or second person paradigm.

\section{Guesser}

Words not found in the lexicon are analysed by a separate finite-state
transducer, the guesser.  We developed a simple, extremely compact and
efficient guesser for French.  It is based on the general assumption
that neologisms and uncommon words tend to follow regular inflectional
patterns.

The guesser is thus based on productive endings (like {\em ment} for
adverbs, {\em ible} for adjectives, {\em er} for verbs).  A given
ending may point to various categories, e.g.~{\em er} identifies not
only infinitive verbs but also nouns, due to possible borrowings from
English.  For instance, the ambiguity class for {\em killer} is {\em
[NOUN-SG VERB-INF]}.

These endings belong to the most frequent ending patterns in the
lexicon, where every rare word weights as much as any frequent word.
Endings are not selected according to their frequency in running
texts, because highly frequent words tend to have irregular endings,
as shown by adverbs like {\em jamais}, {\em toujours}, {\em
peut-\^{e}tre}, {\em hier}, {\em souvent} ({\em never}, {\em always},
{\em maybe}\ldots).

Similarly, verb neologisms belong to the regular conjugation paradigm
characterised by the infinitive ending {\em er}, e.g.~{\em
d\'{e}balladuriser}.

With respect to nouns, we first selected productive endings ({\em
iste}, {\em eau}, {\em eur}, {\em rice}\ldots), until we realised a
better choice was to assign a noun tag to all endings, with the
exception of those previously assigned to other classes.  In the
latter case, two situations may arise: either the prefix is shared
between nouns and some other category (such as {\em ment}), or it must
be barred from the list of noun endings (such as {\em aient}, an
inflectional marking of third person plural verbs).  We in fact
introduced some hierarchy into the endings: e.g.~{\em ment} is shared
by adverbs and nouns, while {\em iquement} is assigned to adverbs
only.

Guessing based on endings offers some side advantages: unknown words
often result from alternations, which occur at the beginning of the
word, the rest remaining the same, e.g.~derivational prefixes as in
{\em isra\'{e}lo-jordano-palestinienne} but also oral transcriptions
such as {\em les z'oreilles} ({\em the ears}), with {\em z'} marking
the phonological liaison.  Similarly, spelling errors which account
for many of the unknown words actually affect the ending less than the
internal structure of the word, e.g.~the misspelt verb forms {\em
appellaient, geulait}.  Hyphens used to emphasise a word, e.g.~{\em
har-mo-ni-ser}, also leave endings unaltered.  Those side advantages
do not however operate when the alternation (prefix, spelling error)
applies to a frequent word that does not follow regular ending
patterns.  For instance, the verb {\em construit} and the adverb {\em
tr\`{e}s} are respectively misspelt as {\em constuit} and {\em
tr\'{e}s}, and are not properly recognised.

Generally, the guesser does not recognise words belonging to closed
classes (conjunctions, prepositions, etc.) under the assumption that
closed classes are fully described in the basic lexicon.  A possible
improvement to the guesser would be to incorporate frequent spelling
errors for words that are not otherwise recognised.

\subsection{Testing the guesser}

We extracted, from a corpus of newspaper articles (Lib\'{e}ration), a
list of 13~500 words unknown to the basic lexicon\footnote{On various
large newspaper corpora, an average of 18~\% words are unknown: this
is mostly due to the high frequency of proper nouns.}.  Of those
unknown words, 9385 (i.e.~about 70~\%) are capitalised words, which
are correctly and unambiguously analysed by the guesser as proper
nouns with more than 95~\% accuracy.  Errors are mostly due to foreign
capitalised words which are not proper nouns (such as {\em Eight}) and
onomatopoeia (such as {\em Ooooh}).

The test on the remaining 4000 non-capitalised unknown words is more
interesting.  We randomly selected 800 of these words and ran the
guesser on them.  1192 tags were assigned to those 800 words by the
guesser, which gives an average of 1.5 tags per word.

For 113 words, at least one required tag was missing (118 tags were
missing as a whole, 4 words were lacking more than one tag: they are
misspelt irregular verbs that have not been recognised as such).  This
means that 86~\% of the words got all the required tags from the
guesser.

273 of the 1192 tags were classified as irrelevant.  This concerned
244 words, which means that 70~\% of the words did not get any
irrelevant tags.  Finally, 63~\% of the words got all the required
tags and only those.

If we combine the evaluation on capitalised and non-capitalised words,
85~\% of all unknown words are perfectly tagged by the guesser, and
92~\% get all the necessary tags (with possibly some unwanted ones).

The test on the non-capitalised words was tough enough as we counted
as irrelevant any tag that would be morphologically acceptable on
general grounds, but which is not for a specific word.  For instance,
the misspelt word {\em statisiticiens} is tagged as {\em [ADJ-PL
NOUN-PL]}; we count the {\em ADJ-PL} tag as irrelevant, on the ground
that the underlying correct word {\em statisticiens} is a noun only
(compare with the adjective {\em platoniciens}).

The same occurs with words ending in {\em ement} that are
systematically tagged as {\em [ADV NOUN-SG]}, unless a longer ending
like {\em iquement} is recognised.  This often, but not always, makes
the {\em NOUN-SG} tag irrelevant.

As for missing tags, more than half are adjective tags for words that
are otherwise correctly tagged as nouns or past participles (which
somehow reduces the importance of the error, as the syntactic
distribution of adjectives overlaps with those of nouns and past
participles).

The remaining words that lack at least one tag include misspelt words
belonging to closed classes ({\em come, tr\'{e}s, vavec}) or to
irregular verbs ({\em constuit}), barbarisms resulting from the
omission of blanks ({\em proposde}), or from the adjunction of
superfluous blanks or hyphens ({\em quand-m\^{e}me, so
ci\'{e}t\'{e}}).  We also had a few examples of compound nouns
improperly tagged as singular nouns, e.g.~{\em
rencontres-t\'{e}l\'{e}}, where the plural marking only appears on the
first element of the compound.

Finally, foreign words represent another class of problematic words,
especially if they are not nouns.  We found various English examples
({\em at, born, of, enough, easy}) but also Spanish, e.g.~{\em
levantarse}, and Italian ones, e.g.~{\em palazzi}.

\section{Conclusion}

We have described the tagset, lexicon and guesser that we built for
our French tagger.  In this work, we re-used an existing lexicon.  We
composed this lexicon with finite-state transducers (mapping rules) in
order to produce a new lexical transducer with the new tagset.  The
guesser for words that are not in the lexicon is described in more
detail.  Some test results are given.  The disambiguation itself is
described in \cite{CT95}.

\vspace{5mm}
\begin{acknowledgments}
We want to thank Irene Maxwell and anonymous referees for useful comments.
\end{acknowledgments}

\appendix

\newpage
\section{An example}

This appendix contains an example of a tagged corpus.

\begin{tabbing}
MMMMMMMi \= \kill
Les \> DET-PL \\
travaux \> NOUN-PL \\
devaient- \> VERB-P3PL \\
-ils \> PRON \\
se \> PC \\
d\a'{e}rouler \> VERB-INF \\
en \> PREP \\
s\a'{e}ance \> NOUN-SG \\
pl\a'{e}ni\a`{e}re \> ADJ-SG \\
ou \> CONN \\
en \> PREP \\
commissions \> NOUN-PL \\
? \> PUNCT \\
Les \> DET-PL \\
d\a'{e}l\a'{e}gu\a'{e}s \> NOUN-PL \\
pouvaient- \> VERB-P3PL \\
-ils \> PRON \\
, \> CM \\
comme \> COMME \\
d' \> PREP \\
habitude \> NOUN-SG \\
, \> CM \\
ent\a'{e}riner \> VERB-INF \\
des \> DET-PL \\
r\a'{e}solutions \> NOUN-PL \\
de \> PREP \\
la \> DET-SG \\
direction \> NOUN-SG \\
du \> PREP \\
parti \> NOUN-SG \\
qui \> CONN \\
n' \> NEG \\
avaient \> VAUX-P3PL \\
pas \> ADV \\
\a'{e}t\a'{e} \> VAUX-PAP \\
discut\a'{e}es \> PAP-PL \\
\a`{a} \> PREP \\
la \> DET-SG \\
base \> NOUN-SG \\
? \> PUNCT \\
Ce \> DET-SG \\
prologue \> NOUN-SG \\
d\a'{e}sordonn\a'{e} \> ADJ-SG \\
enfin \> ADV \\
termin\a'{e} \> ADJ-SG \\
, \> CM \\
le \> DET-SG \\
pr\a'{e}sident \> NOUN-SG \\
en \> PREP \\
exercice \> NOUN-SG \\
de \> PREP \\
la \> DET-SG \\
Ligue \> NOUN-SG \\
qui \> CONN \\
, \> CM \\
selon \> PREP \\
la \> DET-SG \\
r\a`{e}gle \> NOUN-SG \\
de \> PREP \\
la \> DET-SG \\
rotation \> NOUN-SG \\
des \> PREP \\
fonctions \> NOUN-PL \\
, \> CM \\
est \> VAUX-P3SG \\
actuellement \> ADV \\
un \> DET-SG \\
Mac\a'{e}donien \> NOUN-SG \\
, \> CM \\
est \> VAUX-P3SG \\
mont\a'{e} \> PAP-SG \\
\a`{a} \> PREP \\
la \> DET-SG \\
tribune \> NOUN-SG \\
pour \> PREP \\
exposer \> VERB-INF \\
la \> DET-SG \\
nouvelle \> ADJ-SG \\
strat\a'{e}gie \> NOUN-SG \\
des \> PREP \\
communistes \> NOUN-PL \\
yougoslaves \> ADJ-PL \\
. \> PUNCT \\
Dans \> PREP \\
un \> DET-SG \\
discours-fleuve \> NOUN-SG \\
qui \> CONN \\
constituait \> VERB-P3SG \\
le \> DET-SG \\
plus \> ADV \\
petit \> ADJ-SG \\
d\a'{e}nominateur \> NOUN-SG \\
commun \> ADJ-SG \\
des \> PREP \\
positions \> NOUN-PL \\
respectives \> ADJ-PL \\
des \> PREP \\
partis \> NOUN-PL \\
des \> PREP \\
six \> NUM \\
r\a'{e}publiques \> NOUN-PL \\
et \> CONN \\
des \> PREP \\
deux \> NUM \\
provinces \> NOUN-PL \\
autonomes \> ADJ-PL \\
, \> CM \\
Mr \> NOUN-SG \\
Milan \> NOUN-INV \\
Pancevski \> NOUN-INV \\
s' \> PC \\
est \> VAUX-P3SG \\
prononc\a'{e} \> PAP-SG \\
pour \> PREP \\
la \> DET-SG \\
libert\a'{e} \> NOUN-SG \\
d' \> PREP \\
association \> NOUN-SG \\
politique \> ADJ-SG \\
( \> PUNCT \\
et \> CONN \\
donc \> ADV \\
l' \> DET-SG \\
abandon \> NOUN-SG \\
du \> PREP \\
monopole \> NOUN-SG \\
de \> PREP \\
la \> DET-SG \\
Ligue \> NOUN-SG \\
) \> PUNCT \\
, \> CM \\
pour \> PREP \\
la \> DET-SG \\
r\a'{e}forme \> NOUN-SG \\
du \> PREP \\
syst\a`{e}me \> NOUN-SG \\
\a'{e}conomique \> ADJ-SG \\
et \> CONN \\
politique \> ADJ-SG \\
, \> CM \\
ainsi \> ADV \\
que \> CONJQUE \\
du \> PREP \\
fonctionnement \> NOUN-SG \\
de \> PREP \\
la \> DET-SG \\
LCY \> NOUN-INV \\
. \> PUNCT \\
C' \> PRON \\
est \> VAUX-P3SG \\
la \> DET-SG \\
seule \> ADJ-SG \\
fa\c{c}on \> NOUN-SG \\
, \> CM \\
a-t- \> VAUX-P3SG \\
-il \> PRON \\
dit \> PAP-SG \\
, \> CM \\
de \> PREP \\
" \> PUNCT \\
pr\a'{e}server \> VERB-INF \\
les \> DET-PL \\
valeurs \> NOUN-PL \\
de \> PREP \\
la \> DET-SG \\
r\a'{e}volution \> NOUN-SG \\
socialiste \> ADJ-SG \\
yougoslave \> ADJ-SG \\
" \> PUNCT \\
. \> PUNCT \\
\end{tabbing}
\end{document}